\documentclass[10pt,twocolumn,letterpaper]{article}

\usepackage[pagenumbers]{cvpr} 

\usepackage{graphicx}
\usepackage{algorithm}
\usepackage{algorithmicx}
\usepackage{amsmath}
\usepackage{amssymb}
\usepackage{algpseudocode}
\usepackage{dsfont}
\usepackage[inline, shortlabels]{enumitem}
\usepackage{booktabs}
\usepackage{subcaption}
\usepackage{tabularx}
\usepackage{multirow}
\usepackage{caption}
\usepackage{arydshln}
\usepackage{placeins}

\usepackage[pagebackref,breaklinks,colorlinks]{hyperref}

\usepackage[capitalize]{cleveref}
\crefname{section}{Sec.}{Secs.}
\Crefname{section}{Section}{Sections}
\Crefname{table}{Table}{Tables}
\crefname{table}{Tab.}{Tabs.}

\def\cvprPaperID{3828} 
\def\confName{CVPR}
\def\confYear{2022}

\begin{document}

\title{Confidence Propagation Cluster: Unleash Full Potential of Object Detectors}

\author{Yichun Shen$^*$ \and Wanli Jiang\thanks{Equal contributions.} \and Zhen Xu \and Rundong Li \and Junghyun Kwon \and Siyi Li\\
NVIDIA\\
{\tt\small \{ashen,williamj,zhenx,davidli,junghyunk,louli\}@nvidia.com}
}
\maketitle
\begin{abstract}
  It's been a long history that most object detection methods obtain objects by using the non-maximum suppression (NMS) and its improved versions like Soft-NMS to remove redundant bounding boxes.
  We challenge those NMS-based methods from three aspects:
  \begin{enumerate*}[1)]
    \item The bounding box with highest confidence value may not be the true positive having the biggest overlap with the ground-truth box.
    \item Not only suppression is required for redundant boxes, but also confidence enhancement is needed for those true positives.
    \item Sorting candidate boxes by confidence values is not necessary so that full parallelism is achievable.
  \end{enumerate*}

  In this paper, inspired by belief propagation (BP), we propose the Confidence Propagation Cluster (CP-Cluster) to replace NMS-based methods, which is fully parallelizable as well as better in accuracy.
  In CP-Cluster, we borrow the message passing mechanism from BP to penalize redundant boxes and enhance true positives simultaneously in an iterative way until convergence.
  We verified the effectiveness of CP-Cluster by applying it to various mainstream detectors such as FasterRCNN, SSD, FCOS, YOLOv3, YOLOv5, Centernet etc.
  Experiments on MS COCO show that our plug and play method, without retraining detectors, is able to steadily improve average mAP of all those state-of-the-art models with a clear margin from 0.3 to 1.9 respectively when compared with NMS-based methods.
\end{abstract}
\section{Introduction}
\label{sec:intro}
\begin{figure*}[h!]
  \centering
  \includegraphics[width=.8\textwidth]{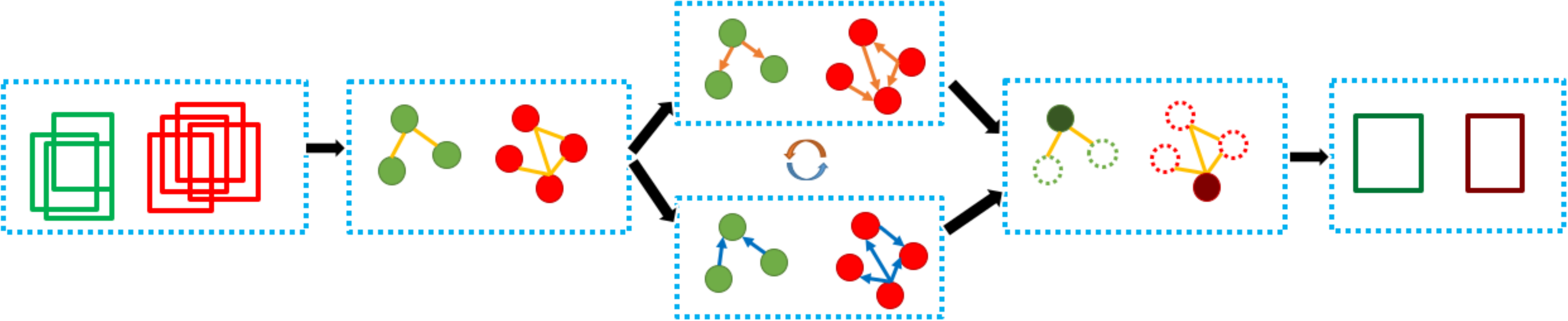}
  \caption{Overall pipeline of CP-Cluster. CP-Cluster converts all candidate boxes from an object detector into a set of graphs. Positive messages (blue arrows) and negative messages (orange arrows) are propagated within each graph iteratively, amplifying true positives and suppressing redundant boxes simultaneously.}
  \label{fig:cp_overall_diagram}
\end{figure*}
The occurrence of convolutional neural networks has brought in revolutionary improvements in various object detection tasks \cite{lin2014microsoft,everingham2015pascal,gupta2019lvis,xia2018dota}.
Generally, two-stage/multi-stage detectors\cite{ren2015faster,dai2016r,cai2018cascade,he2017mask,zhou2021probabilistic} can achieve higher accuracy, while one-stage detectors\cite{liu2016ssd,he2015spatial,lin2017focal,redmon2018yolov3,bochkovskiy2020yolov4,wang2021scaled,21yolov5} take better accuracy-performance balance.
Recently, other than achieving better state-of-the-art results and less inference cost, some research attentions are also paid to simplify training and inference pipelines. \cite{tian2019fcos,law2018cornernet,zhou2019objects,zhang2020bridging} got rid of the predefined anchors. \cite{zhou2019objects, carion2020end, zhu2020deformable, peize2020onenet, wang2021end} designed specific one-one label assignment strategies to train end-to-end detection models without need of post-processing methods.
\cite{chen2021you,zhou2019objects} make use of only one output feature map.

Nowadays some NMS-free methods have gained reasonable accuracies, but they still suffer from more or less sacrifice in accuracies, performance, training time and flexibility of design choices.
Especially, when real-time inference is not required, the ensembles of detection models equipped with NMS are used to achieve better results\cite{zhu2021tph,solovyev2021weighted}. Besides, in autonomous vehicles systems, researchers usually apply NMS to combine objects detected from multiple sensors.
Therefore the majority of those mainstream detectors\cite{21yolov5,lin2017focal,tian2019fcos,ren2015faster} still employ NMS or Soft-NMS\cite{bodla2017soft} to remove redundant bounding boxes in inference stage.
Standard NMS greedily suppresses all neighboring bounding boxes around the box with highest confidence value. Following this, researchers proposed several methods to improve the standard NMS accuracy\cite{bodla2017soft,liu2019adaptive,jiang2018acquisition,zhou2017cad}. Among them, Soft-NMS\cite{bodla2017soft} was proved to achieve general improvements for various detectors, while others are either designed for specific detectors or require retraining with specific tricks.
In addition, some methods are proposed to parallelize the NMS\cite{zheng2021enhancing,bolya2019yolact}, while those methods still rely on confidence sorting in their pipelines.

With those NMS-based methods, all candidate boxes are firstly sorted according to their detection scores, and then the bounding box with the highest score in each cluster is selected as a representative.
Other objects with slightly lower scores are simply thrown away or assigned with a smaller confidence, as does not make full use of relations between candidate boxes.

In this paper, we aim to replace the NMS-based methods with a better clustering framework (CP-Cluster) for object detectors, as is able to achieve better accuracy and meanwhile fully parallelizable.
As illustrated by \cref{fig:cp_overall_diagram}, CP-Cluster firstly constructs a graph set from all candidate boxes based on their overlaps, then both positive messages and negative messages are propagated among boxes belonging to the same graph to tune each box's confidence value until convergence.
In detail, to conquer the deficiencies of NMS-based methods, CP-Cluster is sophisticatedly designed to incorporate below strategies:
\begin{enumerate}[1)]
  \item To make full use of relationships between candidate boxes, we propagate messages among them to tune their confidence values. Specifically, CP-Cluster generates positive messages to enhance true positive boxes and composes negative messages to penalize redundant boxes simultaneously.
  \item To further maximize the confidence margin between true positives and redundant boxes, the confidence message propagations are performed multiple times iteratively.
  \item To achieve full parallelism, the message propagation is restricted within neighboring candidate boxes, so that each candidate box manages to update itself independently.
\end{enumerate}
We summarize our contributions as below:
\begin{enumerate}[1)]
  \item We propose a new fully parallelizable clustering framework (CP-Cluster) applicable for all object detectors which require post-processing, and this new clustering framework outperforms NMS-based methods in accuracies.
  \item We apply CP-Cluster to various mainstream detectors without retraining them, including FasterRCNN\cite{ren2015faster}, SSD\cite{liu2016ssd}, FCOS\cite{tian2019fcos}, yolov5\cite{21yolov5} etc. On MS COCO, experimental results show general improvement for all mainstream detectors by just setting CP-Cluster as post-processing step.
  \item By applying CP-Cluster to CenterNet\cite{zhou2019objects}, we show that some of NMS-free detectors can also be explicitly improved by this clustering framework.
\end{enumerate}

To our knowledge, after Soft-NMS\cite{bodla2017soft}, CP-Cluster is the only bounding box clustering method which manages to achieve general improvements on most of mainstream object detectors in a plug and play manner.
Furthermore, it shows huge potential to be applied in real-time tasks due to its full parallelism.
\section{Related Works}
\label{sec:related_work}
\textbf{Two-stage object detection.} Traditional object detection pipelines mostly employ the sliding window strategy, running a classifier on all ROIs.
Early neural network based methods also follow this way, say the two-stage detectors\cite{girshick2014rich,girshick2015fast,ren2015faster,dai2016r,zhou2021probabilistic}: Candidate ROIs are generated in the first stage, then are further classified in the second stage.
Some subsequent works further improve the accuracy by importing multi-stage detection\cite{cai2018cascade,xiang2017joint}, and \cite{pang2019libra} tries to build relationships between candidate ROIs with RNN. Generally, by employing hierarchical stages, those two-stage methods have the merits of high accuracy, but also suffer from high inference cost and complex training strategies.

\textbf{One-stage object detection.} One-stage detectors\cite{liu2016ssd,fu2017dssd,he2015spatial,lin2017focal,redmon2016you,redmon2017yolo9000,redmon2018yolov3,bochkovskiy2020yolov4,wang2021scaled,tian2019fcos} were proposed with the merits of simpler training pipeline and less inference cost. Some early one-stage detectors were not comparable with two-stage detectors in accuracy, but later works have hugely improved model quality by better training samples selections/assignment strategies\cite{zhang2020bridging,zhu2020autoassign},
stronger neural network architectures\cite{fu2017dssd,redmon2018yolov3,tan2020efficientdet,woo2018cbam}, more sophisticatedly designed loss functions\cite{rezatofighi2019generalized,zheng2020distance,lin2017focal,li2020generalized} and combination of all those techniques\cite{zhang2018single,bochkovskiy2020yolov4,wang2021scaled,21yolov5,tian2019fcos}. Latest methods like YOLO5\cite{21yolov5} have achieved both high accuracy as well as very low inference cost.
One-stage and two-stage detectors are not always competing but can also co-work together as a stronger detector.
For instance, most of those one-stage detectors can be integrated into a two-stage detection framework like FasterRCNN\cite{ren2015faster}, working as the Region Proposal Network\cite{zhou2021probabilistic}.

\textbf{Simplified detectors.} Recently, some research efforts are taken to further simplify the one-stage detectors.
The first direction is to remove predefined anchor boxes during training, simplifying the positive and negative samples assignment strategy\cite{law2018cornernet,tian2019fcos,zhou2019objects,carion2020end,peize2020onenet}.
Secondly, some methods like CenterNet\cite{zhou2019objects} and Yolof\cite{chen2021you} only employ one output feature maps, but still achieve reasonable accuracies. Such simplification may benefit multi-task training, as it allows several tasks to share a same backbone.
Thirdly, starting from keypoint-based detectors\cite{law2018cornernet,zhou2019objects,lan2020saccadenet} and transformer-based detectors\cite{carion2020end,zhu2020deformable}, researchers start to investigate the possibility of end-to-end object detection without post-processing. Specifically, those methods rely on some carefully designed one-one assignment strategies, such as Hungary matching\cite{carion2020end} and minimum cost assignments\cite{peize2020onenet}.

\textbf{Non Maximum Suppression.} Usually a one-one assignment strategy is necessary for an end-to-end detector. However, on the other hand, such strategy restricts detectors from further improving accuracy and reducing inference time cost.
Hence, NMS still works as the most effective post-processing step for the majority of popular object detectors. Other than standard NMS, Soft-NMS \cite{bodla2017soft} assigns lower confidence values for bounding boxes rather than removing them directly, which is more friendly to occlusion case.
\cite{liu2019adaptive} makes use of the density to improve clustering quality specifically for pedestrian detection task. \cite{he2019bounding,jiang2018acquisition} integrate specific tricks into the training progress to co-work with NMS. \cite{learnnms} converted NMS into a learnable neural network. \cite{zheng2020distance} improved NMS by proposing a better overlap computation strategy.
Also, there are some attentions paid on parallelizing the NMS\cite{zheng2021enhancing,bolya2019yolact}, while they still rely on confidence sorting so that they are not fully parallelizable.

\textbf{Belief Propagation in computer vision.} Graph-model based methods have a long history of being applied in computer vision tasks. Some stereo matching tasks\cite{yang2010constant,xiang2012real} make use of BP to smooth the disparity maps.
For scene segmentation tasks, early versions of DeepLab\cite{chen2014semantic} also employ BP as the post-processing step to generate fine-grained segmentation results. Recently, some face clustering methods are fully built upon graph theories to pinpoint face clusters \cite{yang2019learning}

\textbf{Relations to previous methods.} CP-Cluster differentiate from previous NMS-based methods on:
\begin{enumerate}
  \item CP-Cluster is fully built upon graph models and confidence message propagations who no longer follow the framework of NMS.
  \item CP-Cluster is the first bounding box clustering pipeline who tries to enhance true positives and penalize redundant boxes simultaneously.
  \item CP-Cluster does not rely on sorting bounding boxes with confidence values so that full parallelism could be achieved.
\end{enumerate}
Although implemented in different frameworks, CP-Cluster is also compatible with some tricks from previous NMS-based methods: 1) Box coordinates weighting such as \cite{zhou2017cad,solovyev2021weighted}. 2) Different overlaps calculation strategies such as CIOU\cite{zheng2020distance}.
\section{Confidence Propagation Cluster}
\label{sec:cpcluster}
In this section, we discuss details of how CP-Cluster fuses candidate boxes step by step.
We firstly describe how to convert the boxes clustering task to a graph model problem to maximize the confidence margin between true postitives and redundant boxes.
Then we discuss the details of how positive messages and negative messages are composed with heuristics from box distributions to update each candidate box.

\subsection{General Clustering Pipeline}
\label{sec:sub_overview}
\textbf{Building MRFs for bounding boxes.} To describe the neighboring relationships between predicted bounding boxes, we create connections between bounding boxes according to their IOUs and then generalize them into Markov Random Field (MRF) graphs.
For an object detector model, $\mathcal{B}=\{b_1,b_2,b_3,...\}$ is the raw bounding box set from model output before post-processing.
For each box pair $(b_i,b_j\in\mathcal{B})$, we draw an undirected edge between them if their IOU is greater than $\mathbf{\theta}$, generating a set of MRFs $\mathcal{G}=\{g_1,g_2,...\}$.
For each graph $g_i\in\mathcal{G}$, we define $\mathcal{E}_{g_{i}}$ as its edge set and $\mathcal{V}_{g_{i}}$ as its node set.
For a box $b_i\in\mathcal{V}_{g_{n}}$, its neighboring node set $\mathcal{N}_{b_{i}}$ accommodates all nodes connected to $b_i$ in $g_n$.

\begin{figure}[h!]
  \centering
  \begin{subfigure}[b]{0.8\linewidth}
    \label{fig:sub_fig1a}
    \includegraphics[width=\linewidth]{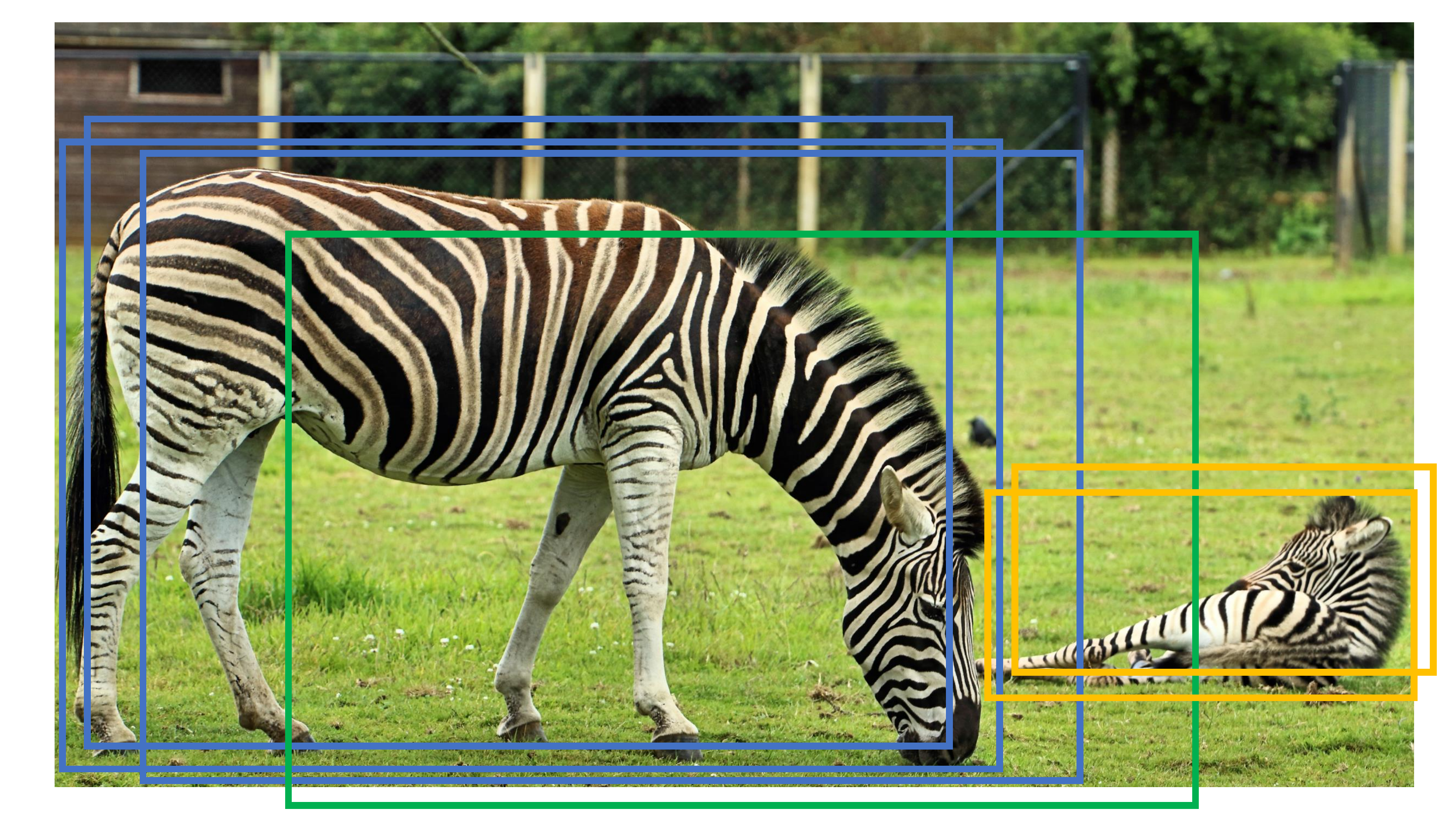}
    \caption{Example of raw detection results before clustering.}
  \end{subfigure}
  \begin{subfigure}[b]{0.8\linewidth}
    \label{fig:sub_fig1b}
    \includegraphics[width=\linewidth]{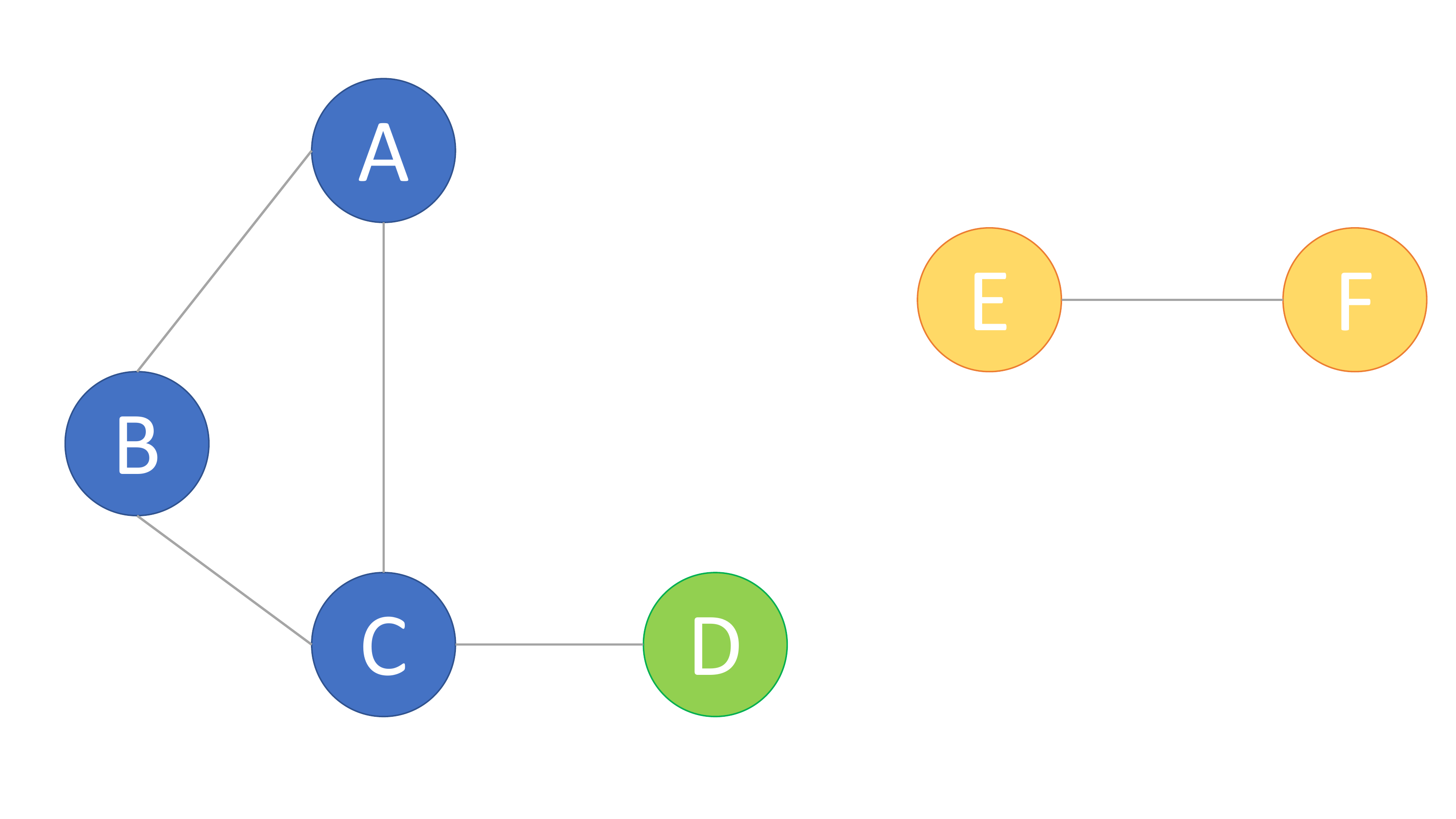}
    \caption{Graph set generated by overlaps of bounding boxes.}
  \end{subfigure}
  \caption{Example of building MRF from bounding boxes by their IOUs($\mathbf{\theta}=0.6$)}
  \label{fig:box_graph}
\end{figure}

\cref{fig:box_graph} is an example of how $\mathcal{G}$ is generated from $\mathcal{B}$ with $\mathbf{\theta}=0.6$, where $\mathcal{B}=\{A,B,C,D,E,F\}$ and $\mathcal{G}=\{g_1,g_2\}$.
In detail, $\mathcal{V}_{g_{1}} = \{A,B,C,D\}$, $\mathcal{E}_{g_{1}} = \{(A,B),(B,C),(A,C),(C,D)\}$, $\mathcal{V}_{g_{2}} = \{E,F\}$, $\mathcal{E}_{g_{2}} = \{(E,F)\}$.
Taking the box $A\in\mathcal{B}$ for example, its neighboring nodes $\mathcal{N}_{A}$ is $\{B,C\}$.
From \cref{fig:box_graph}, it's noticeable that the number of graphs in $\mathcal{G}$ is same to the number of target boxes, while such equivalence doesn't hold true when two heavily occluded ground-truth boxes have overlap greater than $\theta$.

\textbf{Probabilistic objective.} Given a bounding box $b_i\in{\mathcal{B}}$, we define $\mathbf{\hat{P}}(b_i)=\mathbf{\hat{P}}(b_i|\mathcal{N}_{b_{i}},\overline{b_{i}})$ to be the confidence value of $b_i$ from model output given its neighboring boxes and itself, thus the objective of the clustering process can be defined as:
\begin{align}
  \mathbf{\hat{P}}(b_i) = \mathbf{\hat{P}}(b_i|\mathcal{N}_{b_{i}},\overline{b_{i}}) = \left\{
          \begin{array}{ll}
              1.0 & \quad b_i \in \mathcal{B}_p \\
              0.0 & \quad b_i \in \mathcal{B}_n
          \end{array}
      \right. \label{eq::cluster_objective}
\end{align}
where $\mathcal{B}_p$ stands for the set holding true positive candidate boxes having largest overlaps with ground-truth boxes, and $\mathcal{B}_n$ is the set of redundant bounding boxes. $\overline{b_{i}}$ is the observed confidence of $b_i$ from object detectors.
Equation~\eqref{eq::cluster_objective} is targeted to maximize the confidence values of true positives, and meanwhile minimize the confidence values of redundant boxes.
Compared with the objective of traditional NMS, CP-Cluster's objective is different in three aspects:
\begin{enumerate}
  \item NMS-based clustering methods assume that the box of largest confidence value is always the best choice to be selected, but in Equation~\eqref{eq::cluster_objective} this assumption doesn't hold all the time.
  \item We should not only suppress those redundant boxes, but also need to enhance the confidence value of those true positives.
  \item Each candidate box is only impacted by its neighboring bounding boxes.
\end{enumerate}
\textbf{Clustering pipeline.} In our task, unlike the typical case of belief propagation, neighboring bounding boxes not only smooth each other, but also compete with each other.
Hence, we borrow the idea of iterative message passing from belief propagation but generate the messages by heuristics of bounding box distributions instead of traditional ways in BP such as sum-product or max-product.
Specifically, we design the positive message $\mathbf{M_p}$ to reward those true positives, and negative messages $\mathbf{M_n}$ to penalize those redundant boxes.
Both $\mathbf{M_p}$ and $\mathbf{M_n}$ only update confidence values of the bounding boxes by default.

\begin{algorithm}
	\caption{Confidence Propagation Clustering}
  \label{algo::cpcluster}
	\begin{algorithmic}[1]
    \Require $\mathcal{B}$, $\theta$, $F_{gp}$, $F_{gn}$
		\For {$iteration=1,2,\ldots,N$}
      \State Calculate $\mathcal{G}$ with $\theta$
			\ForAll {$b_i$ in $\mathcal{B}$}
        \State $\mathbf{M_p(i)}$ $\leftarrow$ $F_{gp}(\mathcal{G})$ \Comment{Positive msg in \cref{sec:sub_pos_msg}}
        \State $\mathbf{M_n(i)}$ $\leftarrow$ $F_{gn}(\mathcal{G})$ \Comment{Negative msg in \cref{sec:sub_neg_msg}}
				\State $\mathbf{\hat{P}(b_i)}$ $\leftarrow$ $\mathbf{\hat{P}(b_i)}+\mathbf{M_p(i)}-\mathbf{M_n(i)}$
			\EndFor
      \State $\theta \leftarrow \theta+\lambda$
		\EndFor
	\end{algorithmic}
\end{algorithm}
In \cref{algo::cpcluster}, the graph model construction step (line 2) is similar to overlap matrix calculation step in traditional NMS. $F_{gp}$ is the function to generate positive messages by $\mathcal{G}$ (\cref{sec:sub_pos_msg}), and $F_{gn}$ generates negative messages (\cref{sec:sub_neg_msg}).
Line 8 indicates that $\theta$ will be increased by $\lambda$ in each iteration where $\lambda$ is always positive, leading to incremental IOU threshold during the iterative message passing process.
The motivation behind this incremental overlap threshold is: higher overlap two bounding boxes have, more reasonable one of them should be suppressed more than once.
Furthermore, \cref{algo::cpcluster} is fully parallelizable because the confidence value updating step for each box is completely independent.

\begin{figure}[h!]
  \centering
  \includegraphics[width=\linewidth]{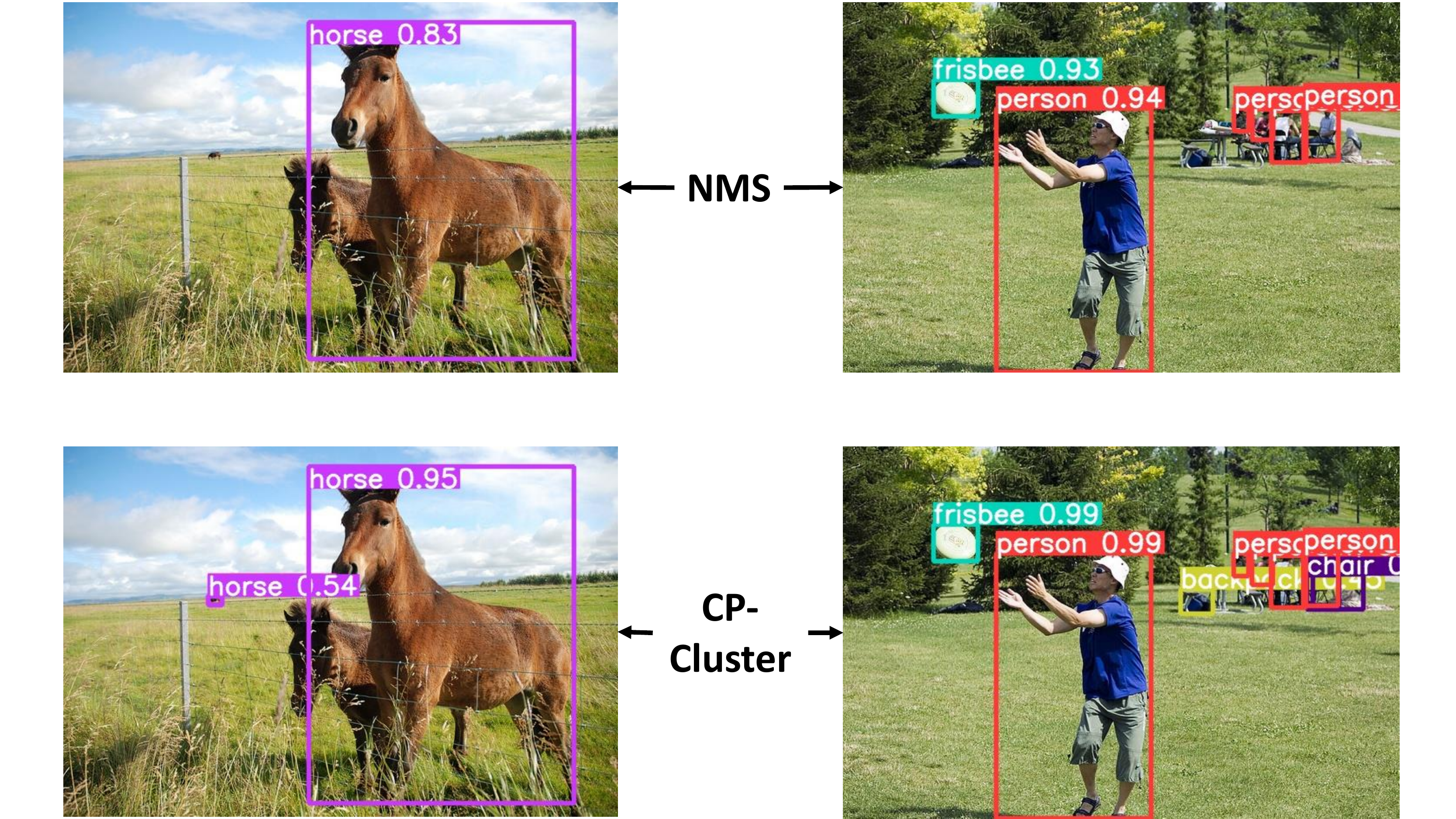}
  \caption{Example of how CP-Cluster enhances true positives and removes redundant boxes in the same time.}
  \label{fig:cp_example}
\end{figure}

\cref{fig:cp_example} is an example of comparison between standard NMS and CP-Cluster. Images in the first row and second row are generated by same Yolov5 model but clustered by standard NMS and CP-Cluster separately and visualized with a constant confidence threshold ($conf>0.4$).
Compared with output boxes from NMS, CP-Cluster not only obtains more objects but also generates higher confidence values for those true positive boxes.

\subsection{Positive Messages Generation}
\label{sec:sub_pos_msg}
One key target of Equation~\eqref{eq::cluster_objective} is to increase the rank of true positive candidate boxes.
For a specific box $b_{i}$, positive messages are generated from its neighboring nodes $\mathcal{N}_{b_{i}}$ to increase $\mathbf{\hat{P}}(b_i)$.

\textbf{Weaker friends aggregation (WFA).} In contrast to high confidence candidate boxes, low confidence boxes are one of the least engaged in traditional post-processing pipelines. As more firewood produce stronger flame, we consider that those low confidence boxes are sometimes evidences to prove their stronger neighbors to be true positives.
With a bounding box $b_i\in\mathcal{V}_{g_{n}}$, his weaker friend set $\mathcal{W}_{b_{i}}$ is a subset of its neighbors $\mathcal{N}_{b_{i}}$, where $IOU(b_{j},b_{i})>\theta_{n}$ and $\mathbf{\hat{P}}(b_j)<\mathbf{\hat{P}}(b_i)$ for each $b_{j}\in\mathcal{W}_{b_{i}}$.
Usually $\theta_{n}$ is greater than the overlap threshold $\theta$ in \cref{algo::cpcluster}, saying that only close enough neighbors can be treated as $b_i$'s friends.
Specifically, we found that the enhancement of a bounding box is mostly affected by below two factors:
\begin{enumerate}
  \item The number of its weaker friends, where such friends indicate stronger enhancement motivation.
  \item The confidence value of its weaker friends. As more friends with high confidence values are evidences to prove that the box itself is true positive. After trying many options, the maximum confidence value of its weaker friends is proved to work best in Equation~\eqref{eq::pmsg_wfa}.
\end{enumerate}
Therefore, we give the definition of positive message generation for a box $b_i$ (Line 4 of \cref{algo::cpcluster}) as below:
\begin{align}
  \mathbf{M_p(i)} \leftarrow \frac{Q}{Q+1} * (1-\mathbf{\hat{P}}(b_i)) * \max \limits_{\hat{b}\in\mathcal{W}_{b_{i}}}\mathbf{\hat{P}}(\hat{b}) \label{eq::pmsg_wfa}
\end{align}
where $Q$ is the number of $b_i$'s weaker friends, and $(1-\mathbf{\hat{P}}(b_i))$ is the normalization term to ensure that the maximum value of $\mathbf{\hat{P}}(b_i)$ won't be greater than 1.0 after applying the positive message.

\begin{figure}[h!]
  \centering
  \begin{subfigure}[b]{0.7\linewidth}
    \label{fig:sub_fig3a}
    \includegraphics[width=\linewidth]{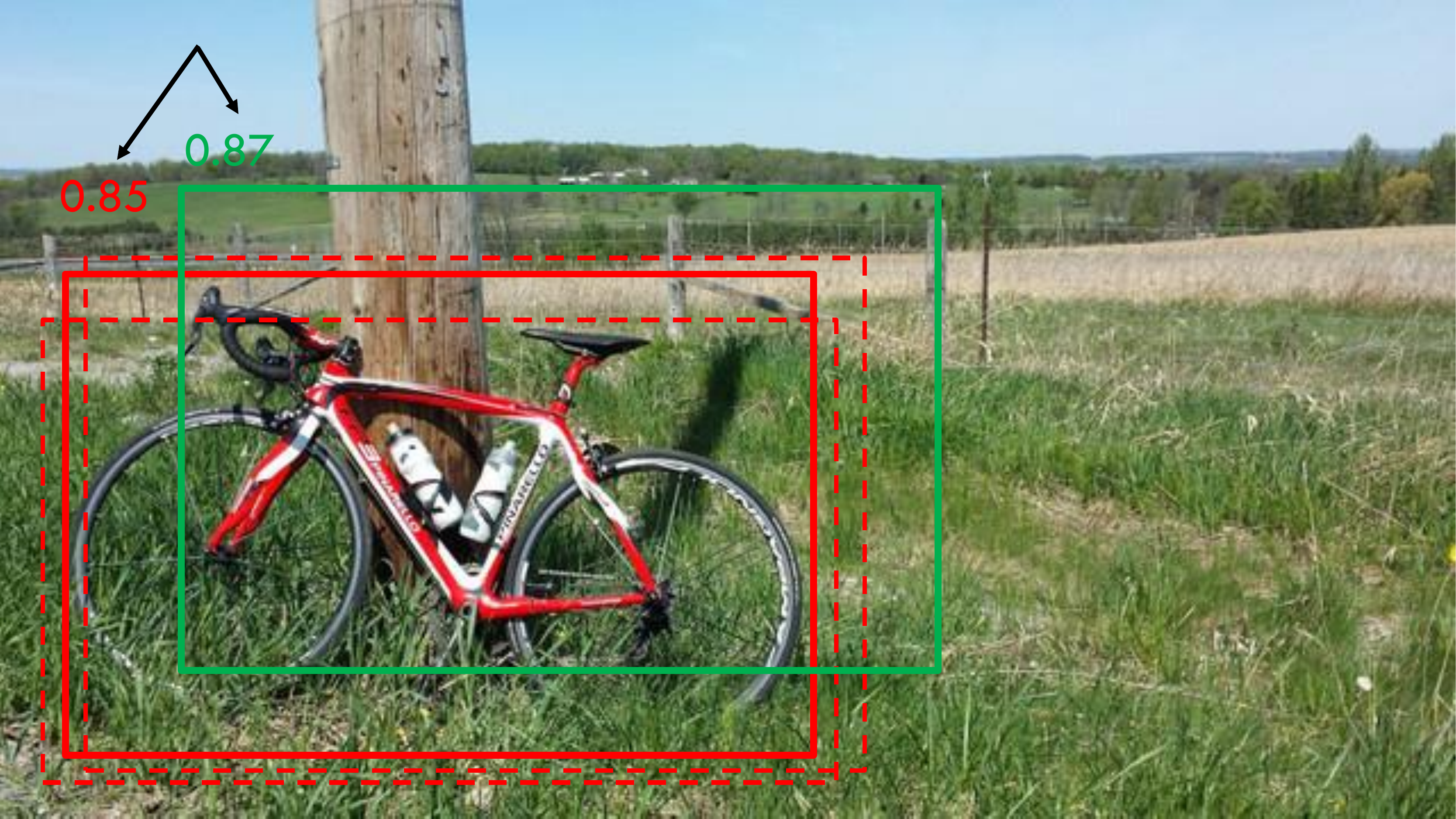}
    \caption{Before WFA.}
  \end{subfigure}
  \begin{subfigure}[b]{0.7\linewidth}
    \label{fig:sub_fig3b}
    \includegraphics[width=\linewidth]{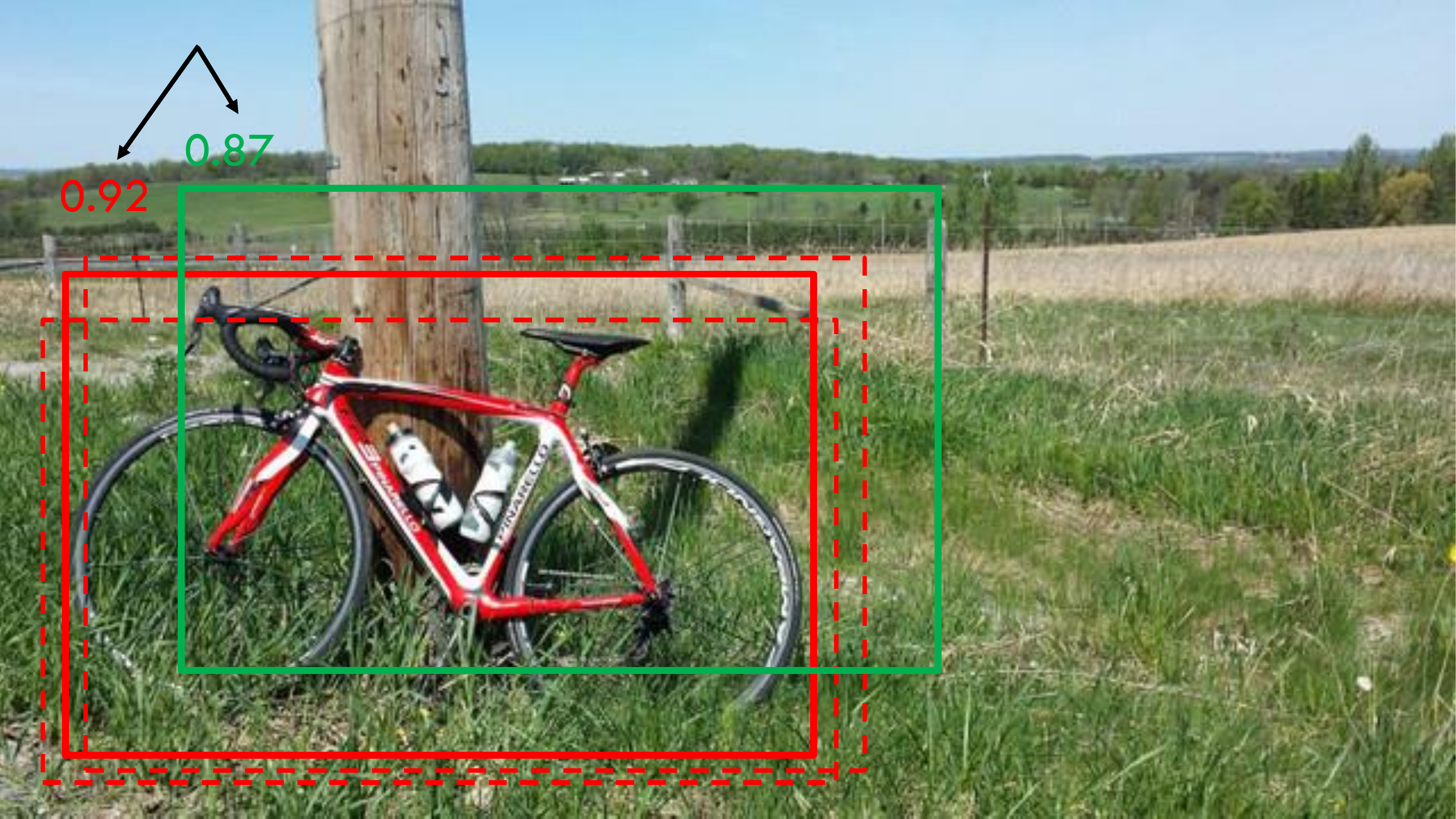}
    \caption{After WFA.}
  \end{subfigure}
  \caption{Confidence value of the bounding box (red solid box) with more weaker friends is enhanced after WFA.}
  \label{fig:wfa_example}
\end{figure}
\cref{fig:wfa_example} is an example on how WFA tune confidence values of bounding boxes. The red solid box is enhanced during the progress of WFA because it has many weaker friends around (red dashed boxes), while the green solid box is not impacted by positive message update due to lack of weaker friends.

\textbf{SNMS-WFA.} To verify the effectiveness of our positive message generation step separately, we integrated the weaker friends aggregation step into standard Soft-NMS, leading to SNMS-WFA.
Specifically, we perform Equation~\eqref{eq::pmsg_wfa} to amplify those important boxes before their weaker friends are suppressed.
In \cref{sec:exp} we will also discuss experimental results of SNMS-WFA and compare it with CP-Cluster.

\subsection{Negative Messages Generation}
\label{sec:sub_neg_msg}
Other than enhancing true positive boxes, suppressing redundant boxes is another objective as indicated by Equation~\eqref{eq::cluster_objective}.\par

Given a bounding box $b_i\in\mathcal{V}_{g_{n}}$, his stronger neighbors $\mathcal{S}_{b_{i}}$ is a subset of $\mathcal{N}_{b_{i}}$, where $IOU(b_{j},b_{i})>\theta$ and $\mathbf{\hat{P}}(b_j)>\mathbf{\hat{P}}(b_i)$ for each $b_{j}\in\mathcal{S}_{b_{i}}$.
In each iteration of \cref{algo::cpcluster}, if a bounding box's stronger neighbor set is not empty, it will be suppressed by one of its stronger neighbor $b_j\in\mathcal{S}_{b_{i}}$.
As to which bounding box is selected to suppress $b_i$, we design the negative impact factor $\mathcal{T}_{(b_{j},b_{i})}$ from $b_j\in\mathcal{S}_{b_{i}}$ to $b_i$ as below:
\begin{align}
  \mathcal{T}_{(b_{j},b_{i})} \leftarrow \alpha*\mathbf{\hat{P}}(b_j)/\mathbf{\hat{P}}(b_i) + (1-\alpha)*IOU(b_j,b_i)/\theta \label{eq::neg_impact}
\end{align}
In Equation~\eqref{eq::neg_impact}, when we set $\alpha=1.0$, the box with largest confidence value in $\mathcal{S}_{b_{i}}$ is selected.
On the contrary, the nearest stronger neighbor with maximum $IOU(b_i,b_j)$ is selected from $\mathcal{S}_{b_{i}}$ if $\alpha=0.0$.

Another problem between $b_i$ and $\mathcal{S}_{b_{i}}$ is worthwhile to be discussed: How many times is a certain box $b_j\in\mathcal{N}_{b_{i}}$ allowed to suppress $b_i$?
For flexibility, we define the suppression counting matrix $\mathbf{SUP}_{j,i}$ to count the times $b_j$ has suppressed $b_i$, and $\zeta$ is the maximum suppression time.
We will discuss more details about how to configure $\zeta$ in \cref{sec:ablation}.

Based on above discussion, the negative message (Line 5 of \cref{algo::cpcluster}) for a box $b_i$ is generated by below equation:
\begin{align}
  \mathbf{M_n(i)} \leftarrow \mathbf{\hat{P}}(b_i) * IOU(b_i,\mathop{\arg\max}\limits_{b_j\in\mathcal{N}_{b_{i}},\mathbf{SUP}_{j,i}<=\zeta}\mathcal{T}_{(b_{j},b_{i})}) \label{eq::nmsg}
\end{align}
Where $\mathbf{SUP}_{j,i}$ is used to restrict the times for $b_i$ to be suppressed by $b_j$, and the box with maximum negative impact factor will be picked up to penalize $b_i$.
\subsection{More Details Behind Confidence Propagation}
\textbf{Message Flow Directions.} As is shown in \cref{sec:sub_pos_msg}, positive messages are passed from weaker boxes to stronger boxes. On the contrary, negative messages flow from stronger boxes to weaker boxes as discussed in \cref{sec:sub_neg_msg}.

\textbf{Parallelism} We have already briefly discussed the parallelism of \cref{algo::cpcluster} in \cref{sec:sub_overview}. Specifically, as each candidate box is only impacted by his neighbors within one iteration, $\mathcal{K}$ threads can be created to handle each box in parallel, where $\mathcal{K}$ is the number of candidate boxes.
Actually, we can further improve the parallelism by combining the graph generation step and message propagation, and $\mathcal{K}$*$\mathcal{K}$ threads can be created to handle the message passing between two boxes.

\section{Experiments}
\label{sec:exp}
\textbf{Dataset.} We conduct experiments on COCO 2017 dataset\cite{lin2014microsoft}. Evaluation results are reported on the COCO val and test-dev dataset.

\textbf{Experiments.} We didn't train new models but directly downloaded models from model zoo for those mainstream detectors.
Then we replace the NMS-based post-process step with CP-Cluster and run the evaluation on COCO val and test-dev dataset.

\textbf{Baselines.} We take standard NMS and Soft-NMS as baselines to compare with our CP-Cluster.
We also performed exhaustive experiments on other plug-and-play NMS versions like weighted-NMS\cite{zhou2017cad} and Clustered-NMS\cite{zheng2021enhancing}, but usually they cannot compete with Soft-NMS or even have negative impacts on some detectors.
For other NMS-based methods like \cite{liu2019adaptive,he2019bounding,learnnms}, they either require retraining models with extra architecture modifications, or are targeted for special tasks.
To save space, we only report baseline metrics for standard NMS and Soft-NMS. In addition, we also report experimental results on WFA-SNMS to prove the effectiveness of our positive message generation strategy separately.

\subsection{Ablation Studies}
\label{sec:ablation}
\begin{figure*}[h!]
  \begin{subfigure}{.33\textwidth}
    \centering
    \includegraphics[width=1.0\linewidth]{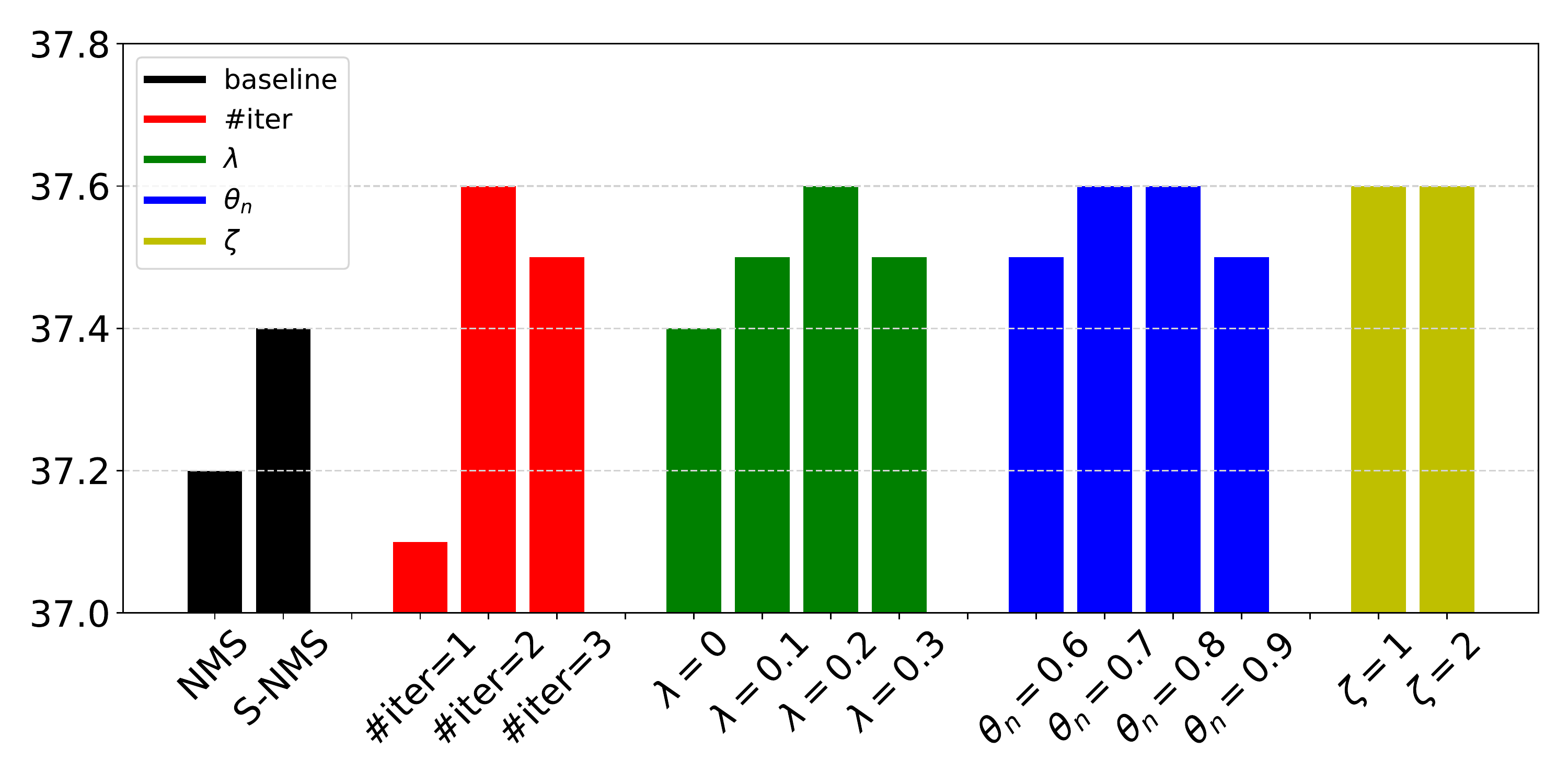}
    \caption{mAP.}
    \label{fig:ablation_map}
  \end{subfigure}
  \hfil
  \begin{subfigure}{.33\textwidth}
    \centering
    \includegraphics[width=1.0\linewidth]{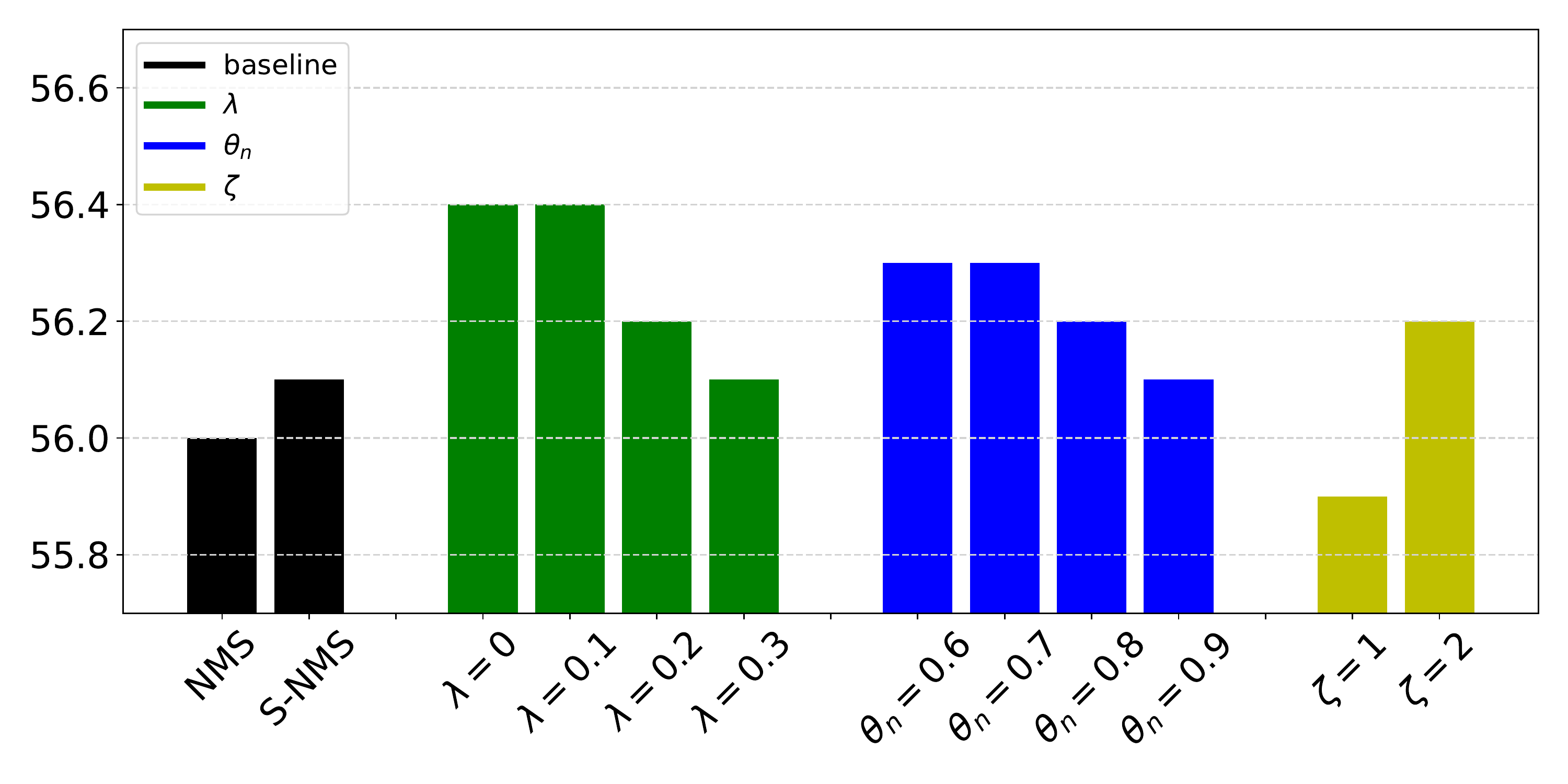}
    \caption{AP50.}
    \label{fig:ablation_ap50}
  \end{subfigure}
  \hfil
  \begin{subfigure}{.33\textwidth}
    \centering
    \includegraphics[width=1.0\linewidth]{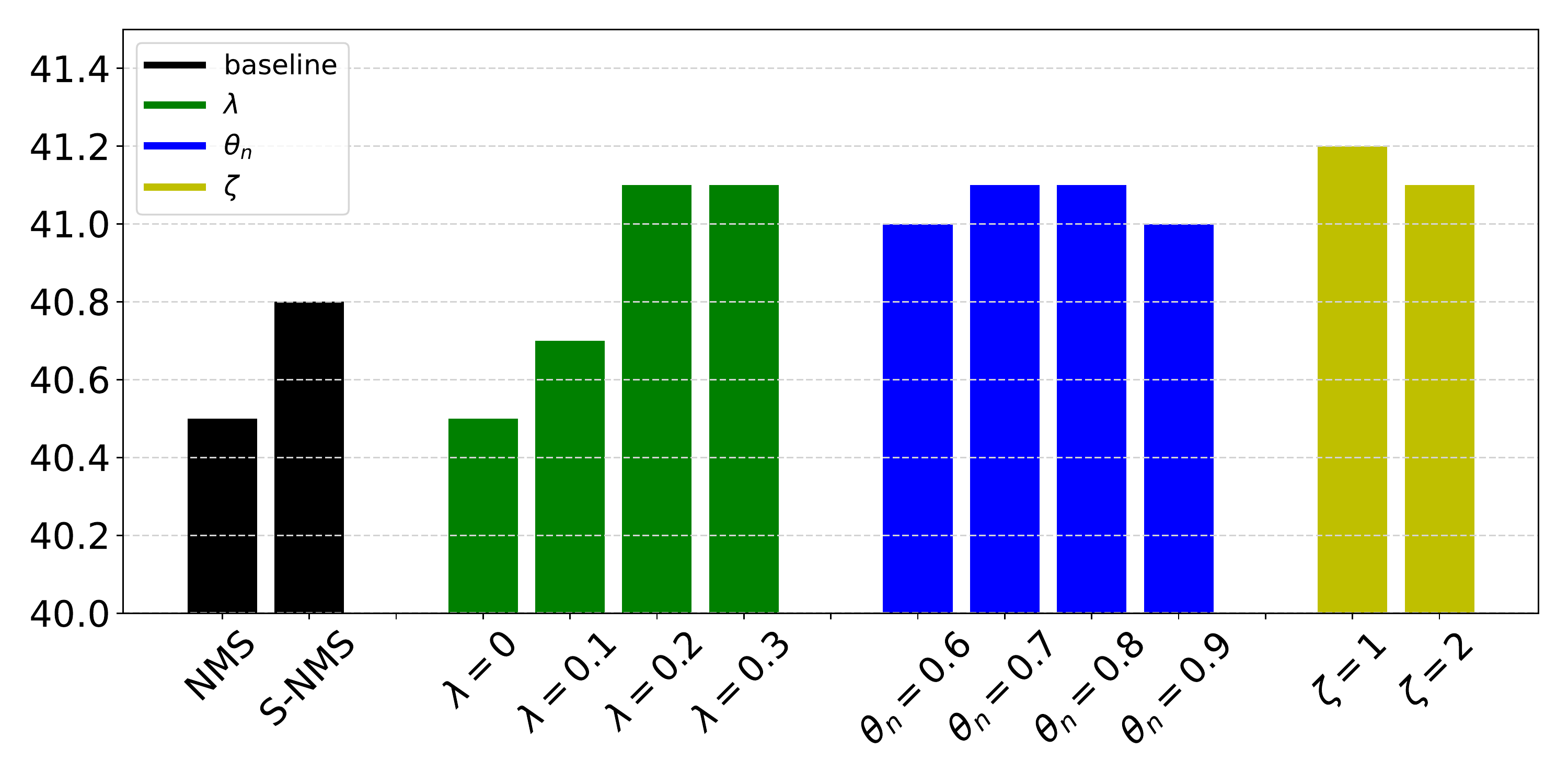}
    \caption{AP75.}
    \label{fig:ablation_ap75}
  \end{subfigure}
  \caption{Accuracies with different hyperparameters on Yolov5s.}
  \label{fig:ablation_figure}
\end{figure*}

All experiments in this section are performed with Yolov5s model downloaded from Yolov5 model zoo. \cref{fig:ablation_figure} shows how mAP, AP50, AP75 are impacted by different hyperparameters separately.

\textbf{Number of iterations.} CP-Cluster provides an iterative way to enhance true positive boxes and meanwhile suppress redundant boxes.
As illustrated by red columns in \cref{fig:ablation_map}, usually $2$ iterations have already been good enough to run the clustering process into convergence.

\textbf{Negative Impact Factor.} In the negative message generation step, negative impact factor is designed to pick up the most appropriate strong neighbor to penalize a box $b_i$ if necessary.
The strong neighbor selection criterion is controlled by the parameter $\alpha$. After trying different options, we found the best result is usually achieved when we apply different $\alpha$ in each iteration.
In detail, we pick up the box with largest confidence value ($\alpha=1.0$) in the first iteration, while in the second iteration we select the box of biggest overlap with $b_i$ ($\alpha=0.0$).

\textbf{Incremental IOU threshold.} From \cref{algo::cpcluster}, the parameter $\lambda$ is used to increment the overlap threshold in each iteration. Intuitively, higher is $\lambda$, less boxes will be penalized in the second iteration.
From green columns in \cref{fig:ablation_figure}, a smaller $\lambda$ leads to better AP50 but worse AP75. In below experiments, we set $\lambda=0.2$ to achieve the most balanced improvements on all buckets.

\textbf{Thresholds to select weaker friends.} In the positive message generation step, the parameter $\theta_{n}$ decides how many boxes are incorporated in the weaker friend set of $b_i$. Specifically, larger $\theta_{n}$ means less friends of $b_i$.
As shown by blue columns in \cref{fig:ablation_figure}, the best accuracy can usually be achieved when $\theta_{n}$ is around $0.8$.

\textbf{Maximum suppression time.} In equation~\eqref{eq::nmsg}, $\zeta$ is used to decide the maximum times a box $b_i$ can be suppressed by $b_j$.
From yellow columns in \cref{fig:ablation_figure}, $\zeta=2$ is beneficial to AP50, while we can get slightly better AP75 when $\zeta=1$. As we found that $\zeta=2$ leads to more stable improvements in most cases, we adopt this setting in our following experiments.

\subsection{Experiments in MMDetection}
\label{sec:sub_exp_mmdetect}
\begin{table*}[h]
  \centering
  \small
  \begin{tabular}{lcccc}
  \toprule
  MAP (val/test-dev)    & nms       & soft-nms  & snms-wfa  & cp-cluster        \\ 
  \midrule
  ssd512           & 29.5/29.6 & 29.8/29.9 & 30.0/30.0 & \textbf{30.1/30.1}      \\ 
  frcnn-r50fpn     & 38.4/38.7 & 39.0/39.2 & 39.1/39.3 & \textbf{39.2/39.4}      \\ 
  fcos-x101        & 42.7/42.8 & 42.7/42.8 & 42.8/42.9 & \textbf{43.0/43.2}      \\ 
  retina-r50fpn    & 37.4/37.7 & 37.5/37.9 & 37.7/38.2 & \textbf{38.1/38.4}      \\ 
  yolov3           & 33.5/33.5 & 33.8/33.8 & 33.6/33.7 & \textbf{34.1/34.1}      \\ 
  yolof            & 37.5/37.8 & 37.6/37.8 & 38.0/38.4 & \textbf{38.1/38.4}      \\ 
  autoassign-fpn50 & 40.4/40.6 & 40.5/40.7 & 40.6/40.8 & \textbf{41.0/41.2}      \\ 
  \bottomrule
  \end{tabular}
  \caption{CP-Cluster with various popular models in MMDetection on COCO val/test-dev.}
  \label{tab:exp_full_mmdetect}
\end{table*}
MMDetection\cite{chen2019mmdetection} is a toolbox with a collection of popular object detector implementations. We implemented our CP-Cluster in mmcv, which is a tool library used by MMDetection.

Since CP-Cluster doesn't require retraining models, we download those popular models from MMDetection model zoo and get them evaluated along with CP-Cluster.
Experimental results are reported on both COCO val and test-dev dataset in \cref{tab:exp_full_mmdetect}.

From \cref{tab:exp_full_mmdetect}, with CP-Cluster, the average mAP of all those popular models are improved by $0.3-0.7$ compared with standard NMS. And compared with Soft-NMS, CP-Cluster still achieved $0.2-0.6$ improvements on average mAP.

\subsection{Experiments With Yolov5}
\label{sec:sub_exp_yolov5}
\begin{table*}[h]
  \centering
  \small
  \begin{tabular}{llccccccc}
  \toprule
  Model                     & Method     & AP            & AP50          & AP75          & APS           & APM           & APL           & AR100         \\ 
  \midrule
  \multirow{2}{*}{s\_640}   & nms        & 37.1          & 55.7          & 40.2          & 20.1          & 41.5          & 45.2          & 55.1          \\
                            & cp-cluster & \textbf{37.4} & \textbf{56.0} & \textbf{40.8} & \textbf{20.3} & \textbf{41.9} & \textbf{45.5} & \textbf{57.2} \\ 
  \hdashline
  \multirow{2}{*}{m\_640}   & nms        & 45.5          & 64.0          & 49.7          & 26.6          & 50.0          & 56.6          & 62.2          \\
                            & cp-cluster & \textbf{45.8} & \textbf{64.2} & \textbf{50.3} & \textbf{26.9} & \textbf{50.3} & \textbf{56.9} & \textbf{64.3} \\ 
  \hdashline
  \multirow{2}{*}{l\_640}   & nms        & 49.0          & 67.3          & 53.4          & 29.9          & 53.4          & 61.3          & 64.6          \\  
                            & cp-cluster & \textbf{49.3} & \textbf{67.4} & \textbf{53.9} & \textbf{30.1} & \textbf{53.7} & \textbf{61.5} & \textbf{67.1} \\ 
  \hdashline
  \multirow{2}{*}{x\_640}   & nms        & 50.7          & 68.8          & 55.1          & 31.9          & 54.9          & 63.4          & 66.6          \\  
                            & cp-cluster & \textbf{51.1} & \textbf{68.9} & \textbf{55.7} & \textbf{32.3} & \textbf{55.2} & \textbf{63.5} & \textbf{68.7} \\ 
  \hdashline
  \multirow{2}{*}{s6\_1280} & nms        & 44.3          & \textbf{62.7} & 48.8          & 27.0          & 48.3          & 53.6          & 62.3          \\  
                            & cp-cluster & \textbf{44.6} & \textbf{62.7} & \textbf{49.4} & \textbf{27.3} & \textbf{48.5} & \textbf{54.1} & \textbf{64.4} \\ 
  \hdashline
  \multirow{2}{*}{m6\_1280} & nms        & 51.2          & \textbf{69.2} & 56.2          & 33.5          & 55.1          & 62.1          & 68.1          \\  
                            & cp-cluster & \textbf{51.5} & \textbf{69.2} & \textbf{56.7} & \textbf{33.7} & \textbf{55.4} & \textbf{62.5} & \textbf{70.2} \\ 
  \hdashline
  \multirow{2}{*}{l6\_1280} & nms        & 53.8          & \textbf{71.6} & 58.9          & 36.3          & 57.8          & 64.9          & 70.3          \\  
                            & cp-cluster & \textbf{54.1} & \textbf{71.6} & \textbf{59.4} & \textbf{36.6} & \textbf{58.1} & \textbf{65.3} & \textbf{72.4} \\ 
  \hdashline
  \multirow{2}{*}{x6\_1280} & nms        & 55.1          & \textbf{72.8} & 60.4          & 37.8          & 58.9          & 66.5          & 71.5          \\  
                            & cp-cluster & \textbf{55.5} & \textbf{72.8} & \textbf{60.9} & \textbf{38.1} & \textbf{59.3} & \textbf{66.8} & \textbf{73.4} \\ 
  \bottomrule
  \end{tabular}
  \caption{CP-Cluster with 8 yolov5 models on COCO test-dev.}
  \label{tab:exp_yolov5}
\end{table*}
Recently Yolov5\cite{21yolov5} is getting popular due to its extreme balance in accuracy and time cost.

In our experiments, we download the pretrained checkpoints (v6 on 1/10/2022) and pair them with our CP-Cluster. For default NMS, we reproduce the evaluation result on COCO test-dev with suggested IOU threshold $\theta=0.65$. While for CP-Cluster, we employ a slightly smaller $\theta=0.6$.

Experimental results are reported on COCO test-dev dataset in \cref{tab:exp_yolov5}, which shows that CP-Cluster manages to achieve $0.3-0.4$ improvements on average mAP compared with standard NMS.
To save table size, we don't report evaluation results for Soft-NMS and SNMS-WFA.
In fact, Soft-NMS fails to make explicitly positive impact on most of Yolov5 models, while SNMS-WFA can achieve similar improvements compared with CP-Cluster.

\subsection{Experiments With Keypoint-based Detectors}
\label{sec:sub_exp_keypoint}
\begin{table*}[ht]
  \centering
  \small
  \begin{tabular}{llccccccc}
  \toprule
  Model                                                                          & Method     & AP            & AP50          & AP75          & APS           & APM           & APL           & AR100         \\ 
  \midrule
  \multirow{3}{*}{dla34}                                                         & maxpool    & 37.3          & 55.1          & 40.7          & 18.6          & 41.1          & 49.2          & 55.8          \\  
                                                                                 & soft-nms   & 38.1          & 57.0          & 41.1          & 18.7          & 40.8          & 50.7          & 56.8          \\  
                                                                                 & cp-cluster & \textbf{39.2} & \textbf{57.9} & \textbf{43.0} & \textbf{20.4} & \textbf{42.4} & \textbf{51.3} & \textbf{58.0} \\ 
  \hdashline
  \multirow{3}{*}{\begin{tabular}[c]{@{}l@{}}dla34\_\\ flip\_scale\end{tabular}} & maxpool    & 41.7          & 60.6          & 45.1          & 21.7          & 44.0          & 56.0          & 60.4          \\  
                                                                                 & soft-nms   & 40.6          & 58.7          & 43.8          & 21.2          & 43.1          & 54.8          & 57.4          \\  
                                                                                 & cp-cluster & \textbf{43.3} & \textbf{61.8} & \textbf{47.6} & \textbf{24.3} & \textbf{45.9} & \textbf{56.4} & \textbf{62.7} \\ 
  \hdashline
  \multirow{3}{*}{hg104}                                                         & maxpool    & 40.2          & 59.1          & 43.8          & 22.5          & 43.4          & 50.8          & 56.0          \\  
                                                                                 & soft-nms   & 40.6          & 58.7          & 44.5          & 23.1          & 43.9          & \textbf{51.0} & 57.4          \\  
                                                                                 & cp-cluster & \textbf{41.1} & \textbf{59.9} & \textbf{45.0} & \textbf{24.4} & \textbf{44.6} & \textbf{51.0} & \textbf{58.4} \\ 
  \hdashline
  \multirow{3}{*}{\begin{tabular}[c]{@{}l@{}}hg104\_\\ flip\_scale\end{tabular}} & maxpool    & 45.2          & 64.1          & 49.3          & 26.7          & 47.2          & 57.9          & 63.2          \\  
                                                                                 & soft-nms   & 44.3          & 62.8          & 48.3          & 26.2          & 46.5          & 57.0          & 60.8          \\  
                                                                                 & cp-cluster & \textbf{46.6} & \textbf{65.0} & \textbf{51.5} & \textbf{28.9} & \textbf{49.0} & \textbf{58.3} & \textbf{65.1} \\ 
  \bottomrule
  \end{tabular}
  \caption{CP-Cluster for Centernet on COCO test-dev.}
  \label{tab:exp_centernet}
\end{table*}
Keypoint-based object detectors\cite{law2018cornernet,zhou2019objects,lan2020saccadenet} are among the earliest attempts to remove the NMS post-process step. Specifically, they replaced the NMS with a simple maxpooling operation to pick up peak points in predicted heatmaps.
As discussed in \cite{zhou2019objects}, NMS methods show positive impacts for some Centernet models but lead to negative results for others.

In our experiments, we download the pretrained models directly from official Centernet repo\cite{zhou2019objects}. For those non-maxpooling based experiments, the maxpooling step is replaced by Soft-NMS and CP-Cluster respectively with IOU threshold $\theta=0.5$. Experimental results on COCO test-dev are reported in \cref{tab:exp_centernet}, where ``dla34\_flip\_scale'' means the model with ``dla34'' arch, augmented by rescaling and flipping.

Compared with default maxpooling post-processing step, all Centernet models are improved with a margin $0.6-1.9$ on average mAP when paired with CP-Cluster, including those models with multi-scale and flip augmentations.
Furthermore, Soft-NMS method can also improve the accuracy of Centernet when they replaced maxpooling in those experiments on single models, while it has negative impacts in multi-scale fusion experiments.
The stable improvements provided by CP-Cluster on multi-scale tests show its potential as a better cluster to handle bounding boxes from multiple models.

\subsection{Experiments for Instance Segmentation}
\begin{table*}[!ht]
  \centering
  \small
  \begin{tabular}{lcccccc}
  \toprule
                     & \multicolumn{2}{c}{NMS}              & \multicolumn{2}{c}{Soft-NMS}                      & \multicolumn{2}{c}{CP-Cluster}                    \\ 
                     & Box AP & Mask AP & Box AP & Mask AP & Box AP & Mask AP \\ 
  \midrule
  MaskRCNN\_R50\_3X  & 41.5   & 37.7    & 42.0          & 37.8          & \textbf{42.2} & \textbf{38.1} \\ 
  MaskRCNN\_R101\_3X & 43.1   & 38.8    & 43.6          & 39.0          & \textbf{43.7} & \textbf{39.2} \\ 
  MaskRCNN\_X101\_3X & 44.6   & 40.0    & \textbf{45.2} & \textbf{40.2} & \textbf{45.2} & \textbf{40.2} \\ 
  \bottomrule
  \end{tabular}
  \caption{CP-Cluster for MaskRCNN on COCO test-dev.}
  \label{tab:exp_maskRCNN}
\end{table*}
Instance segmentation methods are usually built upon object detectors to gain accurate instance area for detected objects.
Still with MMDetection, we apply CP-Cluster to various MaskRCNN models from model zoo, and experimental results on COCO test-dev are shown in \cref{tab:exp_maskRCNN}.
Compared with standard NMS, CP-Cluster shows considerable improvements on both BOX-AP as well as MASK-AP. Although Soft-NMS and CP-Cluster achieve similar accuracy on the X101 model, CP-Cluster outperforms Soft-NMS on all other more lightweight MaskRCNN models.

\subsection{Runtime Measurements}
\label{sec:sub_exp_runtime}
We measure the runtime cost for both CPU and GPU versions of CP-Cluster along with Yolov5 framework. CP-Cluster is compared with CPU Soft-NMS in mmcv and GPU NMS in torchvision.
Note that CP-Cluster does not rely on sorting bounding boxes by their confidence values. However, to make the APIs consistent with torchvision, an extra box sorting step is appended at the end of our CP-Cluster to make sure that true positive boxes are returned in descending order by their confidence values.
When measuring runtime of CP-Cluster on GPU, we exclude the step of box sorting. The measurements are run on a workstation with a 9th-Gen Core-i7 CPU and a Titan-V GPU.

As shown in \cref{tab:exp_runtime}, our GPU implementation of CP-Cluster($Iter=2$) is comparable to the NMS implementation in torchvision. Actually, we are still working on further optimizing the GPU implementation as it will benefit from more sophisticatedly designed CUDA tricks.
\begin{table}[]
  \small
  \begin{tabular}{lccccc}
  \toprule
  Runtime(ms) & NMS & Soft-NMS & \multicolumn{3}{c}{CP(Iter=1,2,3)}               \\
  \midrule
  CPU(mmcv)   & N/A & 11.1     & \multicolumn{1}{c}{32}  & \multicolumn{1}{c}{52}  & 63  \\
  GPU         & 1.4 & N/A      & \multicolumn{1}{c}{1.0} & \multicolumn{1}{c}{1.3} & 1.5 \\
  \bottomrule
  \end{tabular}
  \caption{Runtime Comparison of CP-Cluster.}
  \label{tab:exp_runtime}
\end{table}
\section{Conclusion}
\label{sec:conclusion}
In this work, we have presented a new graph model based bounding box clustering framework (\textbf{CP-Cluster}), which is fully parallelizable.
This framework can work as a general post-processing step for all object detectors, replacing traditional NMS-based methods.
Compared with NMS and Soft-NMS, CP-Cluster is able to achieve better accuracy on MS COCO dataset when applied to the same model.\par

{\small
\bibliographystyle{ieee_fullname}
\bibliography{bp_cluster_bib.bib}

\begin{thebibliography}{10}\itemsep=-1pt

\bibitem{bochkovskiy2020yolov4}
Alexey Bochkovskiy, Chien-Yao Wang, and Hong-Yuan~Mark Liao.
\newblock Yolov4: Optimal speed and accuracy of object detection.
\newblock {\em arXiv preprint arXiv:2004.10934}, 2020.

\bibitem{bodla2017soft}
Navaneeth Bodla, Bharat Singh, Rama Chellappa, and Larry~S Davis.
\newblock Soft-nms--improving object detection with one line of code.
\newblock In {\em Proceedings of the IEEE international conference on computer
  vision}, pages 5561--5569, 2017.

\bibitem{bolya2019yolact}
Daniel Bolya, Chong Zhou, Fanyi Xiao, and Yong~Jae Lee.
\newblock Yolact: Real-time instance segmentation.
\newblock In {\em Proceedings of the IEEE/CVF International Conference on
  Computer Vision}, pages 9157--9166, 2019.

\bibitem{cai2018cascade}
Zhaowei Cai and Nuno Vasconcelos.
\newblock Cascade r-cnn: Delving into high quality object detection.
\newblock In {\em Proceedings of the IEEE conference on computer vision and
  pattern recognition}, pages 6154--6162, 2018.

\bibitem{carion2020end}
Nicolas Carion, Francisco Massa, Gabriel Synnaeve, Nicolas Usunier, Alexander
  Kirillov, and Sergey Zagoruyko.
\newblock End-to-end object detection with transformers.
\newblock In {\em European Conference on Computer Vision}, pages 213--229.
  Springer, 2020.

\bibitem{chen2019mmdetection}
Kai Chen, Jiaqi Wang, Jiangmiao Pang, Yuhang Cao, Yu Xiong, Xiaoxiao Li,
  Shuyang Sun, Wansen Feng, Ziwei Liu, Jiarui Xu, et~al.
\newblock Mmdetection: Open mmlab detection toolbox and benchmark.
\newblock {\em arXiv preprint arXiv:1906.07155}, 2019.

\bibitem{chen2014semantic}
Liang-Chieh Chen, George Papandreou, Iasonas Kokkinos, Kevin Murphy, and Alan~L
  Yuille.
\newblock Semantic image segmentation with deep convolutional nets and fully
  connected crfs.
\newblock {\em arXiv preprint arXiv:1412.7062}, 2014.

\bibitem{chen2021you}
Qiang Chen, Yingming Wang, Tong Yang, Xiangyu Zhang, Jian Cheng, and Jian Sun.
\newblock You only look one-level feature.
\newblock In {\em Proceedings of the IEEE/CVF Conference on Computer Vision and
  Pattern Recognition}, pages 13039--13048, 2021.

\bibitem{dai2016r}
Jifeng Dai, Yi Li, Kaiming He, and Jian Sun.
\newblock R-fcn: Object detection via region-based fully convolutional
  networks.
\newblock In {\em Advances in neural information processing systems}, pages
  379--387, 2016.

\bibitem{everingham2015pascal}
Mark Everingham, SM~Ali Eslami, Luc Van~Gool, Christopher~KI Williams, John
  Winn, and Andrew Zisserman.
\newblock The pascal visual object classes challenge: A retrospective.
\newblock {\em International journal of computer vision}, 111(1):98--136, 2015.

\bibitem{fu2017dssd}
Cheng-Yang Fu, Wei Liu, Ananth Ranga, Ambrish Tyagi, and Alexander~C Berg.
\newblock Dssd: Deconvolutional single shot detector.
\newblock {\em arXiv preprint arXiv:1701.06659}, 2017.

\bibitem{girshick2015fast}
Ross Girshick.
\newblock Fast r-cnn.
\newblock In {\em Proceedings of the IEEE international conference on computer
  vision}, pages 1440--1448, 2015.

\bibitem{girshick2014rich}
Ross Girshick, Jeff Donahue, Trevor Darrell, and Jitendra Malik.
\newblock Rich feature hierarchies for accurate object detection and semantic
  segmentation.
\newblock In {\em Proceedings of the IEEE conference on computer vision and
  pattern recognition}, pages 580--587, 2014.

\bibitem{gupta2019lvis}
Agrim Gupta, Piotr Dollar, and Ross Girshick.
\newblock Lvis: A dataset for large vocabulary instance segmentation.
\newblock In {\em Proceedings of the IEEE/CVF Conference on Computer Vision and
  Pattern Recognition}, pages 5356--5364, 2019.

\bibitem{he2017mask}
Kaiming He, Georgia Gkioxari, Piotr Doll{\'a}r, and Ross Girshick.
\newblock Mask r-cnn.
\newblock In {\em Proceedings of the IEEE international conference on computer
  vision}, pages 2961--2969, 2017.

\bibitem{he2015spatial}
Kaiming He, Xiangyu Zhang, Shaoqing Ren, and Jian Sun.
\newblock Spatial pyramid pooling in deep convolutional networks for visual
  recognition.
\newblock {\em IEEE transactions on pattern analysis and machine intelligence},
  37(9):1904--1916, 2015.

\bibitem{he2019bounding}
Yihui He, Chenchen Zhu, Jianren Wang, Marios Savvides, and Xiangyu Zhang.
\newblock Bounding box regression with uncertainty for accurate object
  detection.
\newblock In {\em Proceedings of the ieee/cvf conference on computer vision and
  pattern recognition}, pages 2888--2897, 2019.

\bibitem{learnnms}
Jan Hosang, Rodrigo Benenson, and Bernt Schiele.
\newblock Learning non-maximum suppression.
\newblock In {\em Proceedings of the IEEE conference on computer vision and
  pattern recognition}, pages 6469--6477, 2017.

\bibitem{jiang2018acquisition}
Borui Jiang, Ruixuan Luo, Jiayuan Mao, Tete Xiao, and Yuning Jiang.
\newblock Acquisition of localization confidence for accurate object detection.
\newblock In {\em Proceedings of the European conference on computer vision
  (ECCV)}, pages 784--799, 2018.

\bibitem{lan2020saccadenet}
Shiyi Lan, Zhou Ren, Yi Wu, Larry~S Davis, and Gang Hua.
\newblock Saccadenet: A fast and accurate object detector.
\newblock In {\em Proceedings of the IEEE/CVF Conference on Computer Vision and
  Pattern Recognition}, pages 10397--10406, 2020.

\bibitem{law2018cornernet}
Hei Law and Jia Deng.
\newblock Cornernet: Detecting objects as paired keypoints.
\newblock In {\em Proceedings of the European conference on computer vision
  (ECCV)}, pages 734--750, 2018.

\bibitem{li2020generalized}
Xiang Li, Wenhai Wang, Lijun Wu, Shuo Chen, Xiaolin Hu, Jun Li, Jinhui Tang,
  and Jian Yang.
\newblock Generalized focal loss: Learning qualified and distributed bounding
  boxes for dense object detection.
\newblock {\em Advances in Neural Information Processing Systems},
  33:21002--21012, 2020.

\bibitem{lin2017focal}
Tsung-Yi Lin, Priya Goyal, Ross Girshick, Kaiming He, and Piotr Doll{\'a}r.
\newblock Focal loss for dense object detection.
\newblock In {\em Proceedings of the IEEE international conference on computer
  vision}, pages 2980--2988, 2017.

\bibitem{lin2014microsoft}
Tsung-Yi Lin, Michael Maire, Serge Belongie, James Hays, Pietro Perona, Deva
  Ramanan, Piotr Doll{\'a}r, and C~Lawrence Zitnick.
\newblock Microsoft coco: Common objects in context.
\newblock In {\em European conference on computer vision}, pages 740--755.
  Springer, 2014.

\bibitem{liu2019adaptive}
Songtao Liu, Di Huang, and Yunhong Wang.
\newblock Adaptive nms: Refining pedestrian detection in a crowd.
\newblock In {\em Proceedings of the IEEE/CVF Conference on Computer Vision and
  Pattern Recognition}, pages 6459--6468, 2019.

\bibitem{liu2016ssd}
Wei Liu, Dragomir Anguelov, Dumitru Erhan, Christian Szegedy, Scott Reed,
  Cheng-Yang Fu, and Alexander~C Berg.
\newblock Ssd: Single shot multibox detector.
\newblock In {\em European conference on computer vision}, pages 21--37.
  Springer, 2016.

\bibitem{pang2019libra}
Jiangmiao Pang, Kai Chen, Jianping Shi, Huajun Feng, Wanli Ouyang, and Dahua
  Lin.
\newblock Libra r-cnn: Towards balanced learning for object detection.
\newblock In {\em Proceedings of the IEEE/CVF Conference on Computer Vision and
  Pattern Recognition}, pages 821--830, 2019.

\bibitem{redmon2016you}
Joseph Redmon, Santosh Divvala, Ross Girshick, and Ali Farhadi.
\newblock You only look once: Unified, real-time object detection.
\newblock In {\em Proceedings of the IEEE conference on computer vision and
  pattern recognition}, pages 779--788, 2016.

\bibitem{redmon2017yolo9000}
Joseph Redmon and Ali Farhadi.
\newblock Yolo9000: better, faster, stronger.
\newblock In {\em Proceedings of the IEEE conference on computer vision and
  pattern recognition}, pages 7263--7271, 2017.

\bibitem{redmon2018yolov3}
Joseph Redmon and Ali Farhadi.
\newblock Yolov3: An incremental improvement.
\newblock {\em arXiv preprint arXiv:1804.02767}, 2018.

\bibitem{ren2015faster}
Shaoqing Ren, Kaiming He, Ross Girshick, and Jian Sun.
\newblock Faster r-cnn: Towards real-time object detection with region proposal
  networks.
\newblock {\em Advances in neural information processing systems}, 28:91--99,
  2015.

\bibitem{rezatofighi2019generalized}
Hamid Rezatofighi, Nathan Tsoi, JunYoung Gwak, Amir Sadeghian, Ian Reid, and
  Silvio Savarese.
\newblock Generalized intersection over union: A metric and a loss for bounding
  box regression.
\newblock In {\em Proceedings of the IEEE/CVF Conference on Computer Vision and
  Pattern Recognition}, pages 658--666, 2019.

\bibitem{solovyev2021weighted}
Roman Solovyev, Weimin Wang, and Tatiana Gabruseva.
\newblock Weighted boxes fusion: Ensembling boxes from different object
  detection models.
\newblock {\em Image and Vision Computing}, pages 1--6, 2021.

\bibitem{peize2020onenet}
Peize Sun, Yi Jiang, Enze Xie, Wenqi Shao, Zehuan Yuan, Changhu Wang, and Ping
  Luo.
\newblock What makes for end-to-end object detection?
\newblock In {\em Proceedings of the 38th International Conference on Machine
  Learning}, volume 139 of {\em Proceedings of Machine Learning Research},
  pages 9934--9944. PMLR, 2021.

\bibitem{tan2020efficientdet}
Mingxing Tan, Ruoming Pang, and Quoc~V Le.
\newblock Efficientdet: Scalable and efficient object detection.
\newblock In {\em Proceedings of the IEEE/CVF conference on computer vision and
  pattern recognition}, pages 10781--10790, 2020.

\bibitem{tian2019fcos}
Zhi Tian, Chunhua Shen, Hao Chen, and Tong He.
\newblock Fcos: Fully convolutional one-stage object detection.
\newblock In {\em Proceedings of the IEEE/CVF international conference on
  computer vision}, pages 9627--9636, 2019.

\bibitem{21yolov5}
Ultralytics.
\newblock Yolov5.
\newblock 2021.

\bibitem{wang2021scaled}
Chien-Yao Wang, Alexey Bochkovskiy, and Hong-Yuan~Mark Liao.
\newblock Scaled-yolov4: Scaling cross stage partial network.
\newblock In {\em Proceedings of the IEEE/cvf conference on computer vision and
  pattern recognition}, pages 13029--13038, 2021.

\bibitem{wang2021end}
Jianfeng Wang, Lin Song, Zeming Li, Hongbin Sun, Jian Sun, and Nanning Zheng.
\newblock End-to-end object detection with fully convolutional network.
\newblock In {\em Proceedings of the IEEE/CVF Conference on Computer Vision and
  Pattern Recognition}, pages 15849--15858, 2021.

\bibitem{woo2018cbam}
Sanghyun Woo, Jongchan Park, Joon-Young Lee, and In~So Kweon.
\newblock Cbam: Convolutional block attention module.
\newblock In {\em Proceedings of the European conference on computer vision
  (ECCV)}, pages 3--19, 2018.

\bibitem{xia2018dota}
Gui-Song Xia, Xiang Bai, Jian Ding, Zhen Zhu, Serge Belongie, Jiebo Luo, Mihai
  Datcu, Marcello Pelillo, and Liangpei Zhang.
\newblock Dota: A large-scale dataset for object detection in aerial images.
\newblock In {\em Proceedings of the IEEE conference on computer vision and
  pattern recognition}, pages 3974--3983, 2018.

\bibitem{xiang2017joint}
Jia Xiang and Gengming Zhu.
\newblock Joint face detection and facial expression recognition with mtcnn.
\newblock In {\em 2017 4th international conference on information science and
  control engineering (ICISCE)}, pages 424--427. IEEE, 2017.

\bibitem{xiang2012real}
Xueqin Xiang, Mingmin Zhang, Guangxia Li, Yuyong He, and Zhigeng Pan.
\newblock Real-time stereo matching based on fast belief propagation.
\newblock {\em Machine vision and applications}, 23(6):1219--1227, 2012.

\bibitem{yang2019learning}
Lei Yang, Xiaohang Zhan, Dapeng Chen, Junjie Yan, Chen~Change Loy, and Dahua
  Lin.
\newblock Learning to cluster faces on an affinity graph.
\newblock In {\em Proceedings of the IEEE/CVF Conference on Computer Vision and
  Pattern Recognition}, pages 2298--2306, 2019.

\bibitem{yang2010constant}
Qingxiong Yang, Liang Wang, and Narendra Ahuja.
\newblock A constant-space belief propagation algorithm for stereo matching.
\newblock In {\em 2010 IEEE Computer Society Conference on Computer Vision and
  Pattern Recognition}, pages 1458--1465. IEEE, 2010.

\bibitem{zhang2020bridging}
Shifeng Zhang, Cheng Chi, Yongqiang Yao, Zhen Lei, and Stan~Z Li.
\newblock Bridging the gap between anchor-based and anchor-free detection via
  adaptive training sample selection.
\newblock In {\em Proceedings of the IEEE/CVF conference on computer vision and
  pattern recognition}, pages 9759--9768, 2020.

\bibitem{zhang2018single}
Shifeng Zhang, Longyin Wen, Xiao Bian, Zhen Lei, and Stan~Z Li.
\newblock Single-shot refinement neural network for object detection.
\newblock In {\em Proceedings of the IEEE conference on computer vision and
  pattern recognition}, pages 4203--4212, 2018.

\bibitem{zheng2020distance}
Zhaohui Zheng, Ping Wang, Wei Liu, Jinze Li, Rongguang Ye, and Dongwei Ren.
\newblock Distance-iou loss: Faster and better learning for bounding box
  regression.
\newblock In {\em Proceedings of the AAAI Conference on Artificial
  Intelligence}, volume~34, pages 12993--13000, 2020.

\bibitem{zheng2021enhancing}
Zhaohui Zheng, Ping Wang, Dongwei Ren, Wei Liu, Rongguang Ye, Qinghua Hu, and
  Wangmeng Zuo.
\newblock Enhancing geometric factors in model learning and inference for
  object detection and instance segmentation.
\newblock {\em IEEE Transactions on Cybernetics}, 2021.

\bibitem{zhou2017cad}
Huajun Zhou, Zechao Li, Chengcheng Ning, and Jinhui Tang.
\newblock Cad: Scale invariant framework for real-time object detection.
\newblock In {\em Proceedings of the IEEE International Conference on Computer
  Vision Workshops}, pages 760--768, 2017.

\bibitem{zhou2021probabilistic}
Xingyi Zhou, Vladlen Koltun, and Philipp Kr{\"a}henb{\"u}hl.
\newblock Probabilistic two-stage detection.
\newblock {\em arXiv preprint arXiv:2103.07461}, 2021.

\bibitem{zhou2019objects}
Xingyi Zhou, Dequan Wang, and Philipp Kr{\"a}henb{\"u}hl.
\newblock Objects as points.
\newblock {\em arXiv preprint arXiv:1904.07850}, 2019.

\bibitem{zhu2020autoassign}
Benjin Zhu, Jianfeng Wang, Zhengkai Jiang, Fuhang Zong, Songtao Liu, Zeming Li,
  and Jian Sun.
\newblock Autoassign: Differentiable label assignment for dense object
  detection.
\newblock {\em arXiv preprint arXiv:2007.03496}, 2020.

\bibitem{zhu2021tph}
Xingkui Zhu, Shuchang Lyu, Xu Wang, and Qi Zhao.
\newblock Tph-yolov5: Improved yolov5 based on transformer prediction head for
  object detection on drone-captured scenarios.
\newblock In {\em Proceedings of the IEEE/CVF International Conference on
  Computer Vision}, pages 2778--2788, 2021.

\bibitem{zhu2020deformable}
Xizhou Zhu, Weijie Su, Lewei Lu, Bin Li, Xiaogang Wang, and Jifeng Dai.
\newblock Deformable detr: Deformable transformers for end-to-end object
  detection.
\newblock {\em arXiv preprint arXiv:2010.04159}, 2020.

\end{thebibliography}
}

\end{document}